\pdfoutput=1

\documentclass[11pt]{article}

\usepackage[final]{ACL2023}

\usepackage{times}
\usepackage{latexsym}
\usepackage{booktabs}
\usepackage{multirow} 
\usepackage{amssymb}

\usepackage{xcolor}
\usepackage{pifont}
\definecolor{darkgreen}{rgb}{0, 0.7, 0}  
\definecolor{darkred}{rgb}{0.75, 0, 0}    

\newcommand{\xmark}{\textcolor{darkred}{\ding{55}}}   

\usepackage{graphicx}
\usepackage{svg}
\usepackage[T1]{fontenc}

\usepackage[utf8]{inputenc}

\usepackage{microtype}

\usepackage{inconsolata}

\title{Don't Erase, Inform! Detecting and Contextualizing Harmful Language in Cultural Heritage Collections}

\author{{Orfeas {Menis Mastromichalakis}}\textsuperscript{1} {Jason Liartis}\textsuperscript{1} {Kristina Rose}\textsuperscript{2} \\ \textbf{Antoine Isaac}\textsuperscript{3,4} \and \textbf{Giorgos Stamou}\textsuperscript{1} \\
       \textsuperscript{1} National Technical University of Athens  \textsuperscript{2} DFF - Deutsches Filminstitut \& Filmmuseum\\ \textsuperscript{3} Europeana Foundation 
       \textsuperscript{4} Vrije Universiteit Amsterdam \\ 
        \texttt{\{menorf, jliartis\}@ails.ece.ntua.gr, rose@dff.film} \\ \texttt{antoine.isaac@europeana.eu, gstam@cs.ntua.gr  } }

\begin{document}
\maketitle
\begin{abstract}
Cultural Heritage (CH) data hold invaluable knowledge, reflecting the history, traditions, and identities of societies, and shaping our understanding of the past and present. However, many CH collections contain outdated or offensive descriptions that reflect historical biases. CH Institutions (CHIs) face significant challenges in curating these data due to the vast scale and complexity of the task. To address this, we develop an AI-powered tool that detects offensive terms in CH metadata and provides contextual insights into their historical background and contemporary perception. We leverage a multilingual vocabulary co-created with marginalized communities, researchers, and CH professionals, along with traditional NLP techniques and Large Language Models (LLMs). Available as a standalone web app and integrated with major CH platforms, the tool has processed over 7.9 million records, contextualizing the contentious terms detected in their metadata. Rather than erasing these terms, our approach seeks to inform, making biases visible and providing actionable insights for creating more inclusive and accessible CH collections. 
\end{abstract}

\section{Introduction}
Language is central to Cultural Heritage, shaping how professionals and the public interpret historical collections.
However, the descriptions of CH objects can contain outdated or offensive language that reflect historical cultural and societal norms and power structures no longer appropriate today. While CHIs recognize this issue, updating legacy metadata is a daunting task due to the sheer scale of their collections and the complexity of the process.


Artificial Intelligence (AI) offers scalable solutions by performing sophisticated tasks and processing large volumes of data. However, AI systems can inadvertently perpetuate biases inherent in the training data or developers' assumptions. To mitigate this, human-in-the-loop approaches are crucial, integrating input from domain experts, marginalized communities, and CH professionals to ensure both accuracy and cultural sensitivity.

With this work, we aim to improve CH metadata, by making existing bias visible, guiding users in understanding the original context of words, and how they are currently perceived. To do so, we introduce a multilingual vocabulary and an AI-powered tool for detecting offensive terms in CH metadata, providing contextual explanations and suggested alternatives. The vocabulary, co-created with marginalized communities, researchers, and CH professionals, pairs biased language with contextual information and suggestions for the appropriate usage of terms. It is structured as a Knowledge Graph (KG) following the approach of~\citet{nesterov2023knowledge} and standards like the Simple Knowledge Organization System (SKOS)~\cite{baker2013key}. Both the created vocabulary\footnote{\url{https://op.europa.eu/en/web/eu-vocabularies/dataset/-/resource?uri=http://publications.europa.eu/resource/dataset/de-bias-vocabulary}}, and the respective ontology\footnote{\url{https://op.europa.eu/de/web/eu-vocabularies/dataset/-/resource?uri=http://publications.europa.eu/resource/dataset/de-bias-ontology}} describing its structure, are available on EU Vocabularies\footnote{\url{https://op.europa.eu/en/web/eu-vocabularies}}, a platform by the Publications Office of the EU that provides controlled vocabularies and reference data. This publication enhances the vocabulary’s reach and facilitates its integration into broader CH initiatives. 
The system was evaluated through a series of human-in-the-loop validation campaigns refining both the tool and vocabulary. The resulting dataset from these campaigns is published as a benchmark for derogatory term detection in CH.

To ensure accessibility, we developed a web application\footnote{\url{https://debias-tool.ails.ece.ntua.gr/}} that allows manual text input, batch processing, and exploration of results. 
Additionally, we provide an API for integration. Through this API, the tool has been deployed on Europeana\footnote{\url{https://www.europeana.eu/}}, the European digital heritage platform, and CH metadata ingestion platforms like MINT\footnote{\url{https://mint-projects.image.ntua.gr/mint4all}}, processing over 7.9 million records. A dedicated UI component in Europeana now informs users about detected contentious terms. The resources and code used throughout this paper can be found in our GitHub repository\footnote{\url{https://github.com/ails-lab/de-bias}}.

\section{Related Work}

Our work relates to research on toxic, abusive, and offensive language, as well as hate speech. Prior efforts have sought to establish a unified taxonomy of harmful content to enable structured analysis \cite{banko-etal-2020-unified, kurrek-etal-2020-towards}. In this work, we use the umbrella terms hateful or harmful language \cite{hateLab}.

Detecting harmful language is a critical challenge that was traditionally handled manually by moderators, administrators, or curators. However, the vast volume of content now demands automated solutions \cite{waseem-hovy-2016-hateful}. Research in this area spans various domains, including workplace communications \cite{bhat2021say}, online chatbots \cite{song2024detecting}, social media \cite{niraula-etal-2021-offensive}, and news articles \cite{subbiah-etal-2023-towards}.
Early methods relied on SVMs \cite{warner-hirschberg-2012-detecting}, often enhanced with curated dictionaries \cite{tulkens2016dictionary} and ensemble models \cite{burnap2014hate}. Later, deep learning techniques, such as CNNs and LSTMs, became popular \cite{del2017hate, mathur-etal-2018-detecting, meyer-gamback-2019-platform, chakrabarty-etal-2019-pay, modha-etal-2018-filtering}. Recent advances have focused on transformer-based \cite{elmadany-etal-2020-leveraging, alonso2020hate, davidson-etal-2020-developing} and LLM-based approaches \cite{yao2024personalised, plaza2023respectful}.

However, harmful language datasets have been shown to suffer from robustness issues \citep{korre2023harmful} and inherent biases \cite{wiegand2019detection, davidson-etal-2019-racial, xia-etal-2020-demoting, zhang-etal-2020-demographics}. Similarly, data-driven methods often exhibit limitations in robustness \cite{Kaushik2020Learning, sen-etal-2022-counterfactually} and racial biases \cite{sap-etal-2019-risk}, raising concerns about their fairness, real-world applicability, and potential to perpetuate harmful biases \cite{thiago2021fighting}.
Context is essential in detecting harmful language \cite{bourgeade-etal-2024-humans}. However, many detection models struggle to account for context, leading to high false-positive rates, as noted in \cite{kennedy-etal-2020-contextualizing}. Our work addresses this issue through an LLM-based disambiguation module that leverages a curated vocabulary, explicitly defining the context in which a term is considered derogatory. Another key aspect of our approach is the involvement of affected communities in term collection, context definition, and system evaluation. This aligns with recent calls for equitable representation in such tasks \cite{kurrek-etal-2020-towards} and research emphasizing the importance of community context \cite{saleem2022enriching} and cultural nuances in harmful language detection \citep{zhou-etal-2023-cultural}.

The CH domain presents unique challenges that demand a more nuanced approach. Unlike user-generated content, CH metadata references historical entities, events, and societal norms that reflect the language of their time. Simply removing or censoring problematic terms poses risks akin to erasing parts of history and limiting opportunities for critical engagement. Instead, harmful language detection in CH must balance historical accuracy with ethical concerns, ensuring metadata remains both informative and sensitive to evolving societal values. 

In CH, efforts to address contentious terminology have built upon glossaries of problematic terms in museum databases \citep{modest2018words} and evolved into knowledge graphs that encode expert knowledge for more inclusive digital collections \cite{nesterov2023knowledge}. We build on this by extending their proposed KG schema, enriching the vocabulary with additional terms, and expanding coverage beyond English and Dutch.



\section{The Vocabulary}
\label{sec:vocabulary}
The vocabulary is a comprehensive resource designed to identify contentious terms, explain their historical and contextual significance, and guide appropriate usage. It powers the harmful language detection tool (see Section~\ref{sec:tool}), enabling the identification of outdated or potentially harmful language in CH metadata. Additionally, it serves as a standalone reference for CH professionals, and curators to review and refine collection descriptions.

\subsection{Co-creating the Vocabulary}
\label{sec:creating}
The vocabulary includes contentious terms in English, French, German, Dutch, and Italian that are known or likely to appear in CH metadata. It primarily focuses on `Ethno-religious identity', `Gender and sexual identity', and `Migration and colonial past', with extensions to other domains, such as disability. Each term is accompanied by contextual explanations of its problematic nature and recommendations for appropriate handling. 
The vocabulary was developed iteratively, grounded in a typology of bias patterns \cite{masschelein2024bias} that categorizes bias by factors such as religion, sexuality, and ethnicity through an intersectional lens \cite{crenshaw2013mapping}. This framework was combined with an analysis of linguistic patterns, including euphemisms, dysphemisms, and diminutive, that perpetuate bias in CH metadata. 
Editorial guidelines ensured consistency and usability, recommending lower-secondary school reading levels for contextual descriptions, avoiding Eurocentric generalizations, and providing clear recommendations. Additionally, to support UI integration and effectively inform users about contentious terms, we included recommendations on the format and length of descriptions for contentious terms and issues, ensuring compatibility with interfaces that adhere to established UI standards.

The vocabulary compilation followed a twofold approach:
first, we consulted existing glossaries, research publications, encyclopedias, and dictionaries. Scientific literature, along with glossaries, informed contextual descriptions covering etymology, historical usage, and debates around the selected terms. We also examined term occurrences in CH platforms like Europeana to assess their relevance. These sources guided recommendations for handling contentious terms or suggesting alternatives. A full list of the 170+ sources used is available in our GitHub repository\footnote{\url{https://github.com/ails-lab/de-bias/blob/main/resources/vocabulary_sources.csv}}.
Second, terms and descriptions were collected, reviewed, and refined through 12 co-creation sessions, which served as safe spaces for debating and discussing bias issues with affected communities and their allies. 
The sessions were language-specific and focused on distinct topics: (i) Ethnicity and ethno-religious identity with Jewish communities (German), (ii) Gender and sexual identity with LGBTQIA+ communities (Italian, English), and (iii) Migration and colonial history with Dutch- and French-speaking participants. Only terms relevant to the community's thematic focus were discussed in depth during each session. In total, over 60 community members and allies contributed. We refer to the \textit{Ethical Considerations} section at the end of this paper for more details about the support and remuneration of the participants of these events who worked on this sensitive topic. 

To guide and evaluate community engagement, we developed a methodology\footnote{\url{https://pro.europeana.eu/files/Europeana_Professional/Projects/debias/a_community_engagement_methodology_resources_reflections_recommendations_v3_july_2024.pdf}} focused on building and maintaining relationships with communities while ensuring a respectful and inclusive approach. Although specific setups and tools were not predetermined, three primary interaction patterns emerged:
(i) Participants searched CH platforms (e.g. Europeana.eu, Deutsche Digitale Bibliothek\footnote{\url{www.deutsche-digitale-bibliothek.de}}) and institutional databases for contentious terms—either pre-selected by the editorial team or suggested by the community—then debated their findings.
(ii) Existing CH object descriptions were reviewed for accuracy, inclusiveness, and positionality. Discussions explored whether descriptions should retain historical language or take a more interpretive approach to balance objectivity and subjectivity.
(iii) In follow-up meetings, terms and contextual descriptions were revisited for further discussion, allowing for iterative refinement.
These sessions offered valuable insights into community perspectives on contextualization. Participants emphasized that beyond identifying harmful language, CH metadata often suffers from poor representativeness, failing to reflect the viewpoints and knowledge of the communities. Many highlighted the importance of contextualizing contentious terms rather than merely replacing or masking them, as this approach preserves historical accuracy, acknowledges past discriminations, and recognizes the resilience of those who resisted them.

\subsection{Challenges}
Several challenges emerged during the development of the vocabulary. One major challenge was its multilingual nature, as bias does not directly translate across languages. For example, in English, `queer'once had a negative connotation before being reclaimed by the LGBTQIA+ community, whereas in Italian, it was introduced more recently with a solely positive meaning. Consequently, `queer' was included in the English subset but not the Italian one. Similarly, terms with equivalents across languages required language-specific contextualization. 
To address these discrepancies, we developed independent language-specific subsets, enabling editors and community participants to focus on their respective linguistic contexts while optionally cross-referencing other languages. This approach reduced editorial complexity and allowed more thorough research and community engagement. While terms were not forcibly unified across languages at this stage, their semantic relationships are represented in the knowledge graph (see Section~\ref{sec:KG}). For example, `Third World' (English), `Derde Wereld' (Dutch), and `Dritte Welt' (German) are distinct entries linked semantically. Although some links have been established, further refinement will be pursued in future iterations, ensuring an iterative and flexible process without compromising expressivity.
This approach also accommodated imbalances in thematic coverage across languages, stemming from the focus areas of community engagement. For instance, the Dutch vocabulary emphasizes migration and the colonial past, reflecting the priorities of Dutch-speaking participants, while the German subset focuses more on Ethno-religious identity due to its emphasis on antisemitism. Additionally, differences in existing research on diversity, inclusion, and discrimination across countries and languages further contributed to variations in vocabulary size and focus.

Another set of challenges pertained to describing the contentiousness of terms. First, historical accuracy was crucial, as CH metadata often references historical artifacts or language from past decades or centuries. However, meanings evolve over time, requiring descriptions that reflect these shifts while enabling consistent flagging, regardless of data’s creation date, which is often unknown. Second, some terms are used in both contentious and neutral contexts, necessitating careful handling to avoid false positives. Approximately 25\% of the collected terms exhibit such ambiguity. For example, `race' in English can refer to the contested concept of human races or to a competitive event. To address this, we implemented a filtering mechanism in the harmful language detection tool that assesses contextual usage before flagging a term. During vocabulary development, ambiguous terms were explicitly marked, and descriptions were structured to highlight problematic uses while acknowledging neutral ones where necessary. 
By tackling these challenges, the vocabulary maintains linguistic sensitivity, historical accuracy, and contextual precision, ensuring effective detection of harmful terms across diverse CH metadata.

\subsection{Structure and Statistics}
\label{sec:KG}
To ensure interoperability and computational accessibility, we structured our vocabulary as a Knowledge Graph (KG) based on the SKOS standard~\cite{baker2013key}. This design enables a structured representation of terms and their relationships while ensuring scalability and flexibility for future updates. Our approach builds on prior work~\cite{nesterov2023knowledge}, which we extended and adapted to our specific needs.
\begin{figure}
    \centering
    \includegraphics[width=0.35\textwidth]{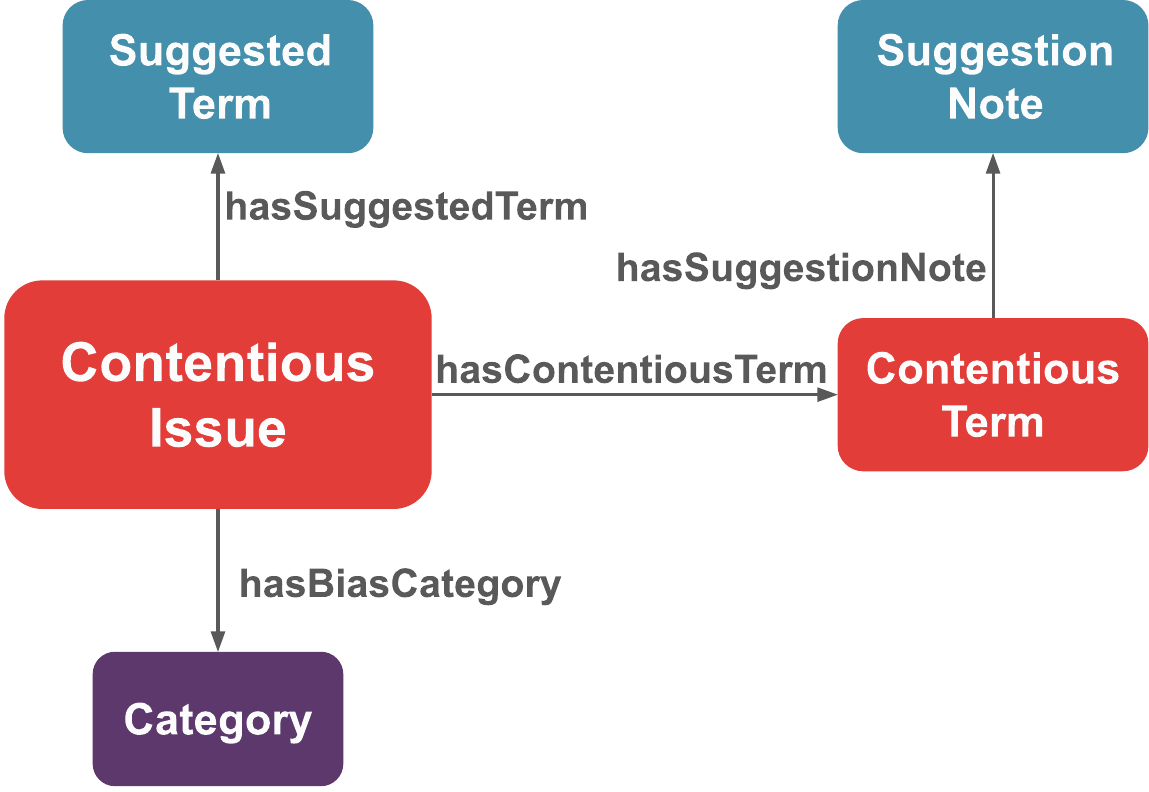} 
    \caption{A simplified version of the KG structure}
    \label{fig:kg}
\end{figure}
At the core of the KG introduced in \citet{nesterov2023knowledge} are contentious terms, their associated contentious issues, suggested alternatives, and suggestion notes.
The ContentiousTerm entity represents each term in the vocabulary, while ContentiousIssue provides contextual information, including etymology, relevant discourse, and references explaining why a term (or set of terms) is considered problematic.SuggestionNotes offer standardized and multilingual recommendations on appropriate usage, and SuggestedTerms provide alternative wording. Synonyms and different terms within the same context are grouped by linking multiple ContentiousTerms to the same ContentiousIssue, while semantic matching across languages is achieved with the appropriate SKOS properties.

We slightly modified the linking between contentious issues, suggestions and labels and extended their structure in several ways. Firstly, we classified the issues using bias categories, as described in Section~\ref{sec:creating}, directly linking them to the relevant contentious issues. To support harmful language detection tools, ambiguous terms are flagged for contextual analysis. We also grouped gendered forms, and spelling variations by linking them to the same ContentiousIssue similarly to how synonyms were treated. Finally, our KG also includes sources and administrative metadata, such as versioning, modification history, and rights information, as a means to foster sustainable reuse of our data by consuming services. Figure~\ref{fig:kg} illustrates a simplified version of our KG structure adapted from~\citet{nesterov2023knowledge}. For further details on the knowledge graph structure, we refer the reader to \citet{debias-report}.

Currently, the vocabulary consists of 687 contentious terms linked to 530 distinct contentious issues. The English subset contains 220 terms, followed by German (163), Dutch (161), French (75), and Italian (68). The vocabulary has been made publicly available through its publication to the EU-Vocabularies, ensuring ongoing maintenance and enhancing its reach and broader adoption. In Appendix \ref{app:voc_example} we provide an example of a vocabulary term.

\section{The tool}
\label{sec:tool}
\begin{figure*}[h]
    \centering
    \includegraphics[width=0.9\textwidth]{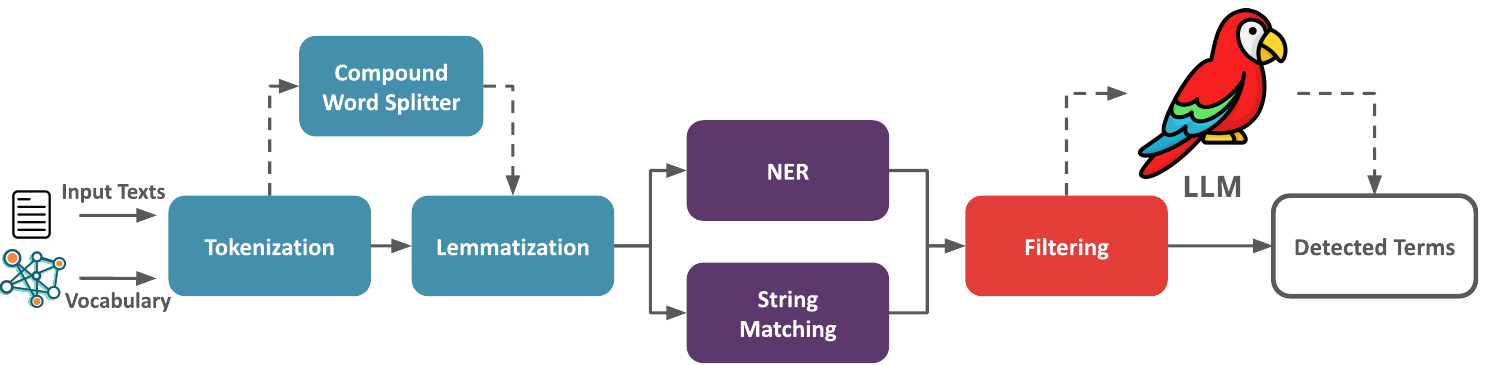} 
    \caption{Tool Architecture Overview}
    \label{fig:tool-overview}
\end{figure*}

Figure~\ref{fig:tool-overview} illustrates the tool's architecture, designed to detect contentious terms leveraging the structured vocabulary described in Section~\ref{sec:vocabulary}. First, inputs undergo tokenization, followed by compound word splitting for German and Dutch, enabling the detection of contentious terms within arbitrarily formed compounds. Next, a lemmatization module standardizes word forms, ensuring accurate matching. The lemmatized texts are then processed independently by two modules: Named Entity Recognition (NER), which identifies proper names, and string matching, which detects vocabulary terms within the input text. These outputs are then merged in a filtering module, which discards detections that are part of named entities.
For terms flagged as ambiguous in the vocabulary, an LLM-based disambiguation module is applied. This module evaluates the surrounding context, alongside the contentious issue description and suggestion notes, to determine whether a term is used in a derogatory manner and should be flagged. The final output consists of detected contentious terms within the input text.
The following subsections provide a detailed breakdown of the main components.

\subsection{Compound Word Splitting}
In German and Dutch, it is common to arbitrarily form compound words, combining multiple terms into a single unit. 
To detect contentious terms in such cases, we developed a compound word splitter, applied before lemmatization to ensure that the lemmatizer processes individual word components.
For German, we adapted an open-source compound splitter\footnote{\url{https://github.com/repodiac/german_compound_splitter/}}, which relies on an external dictionary\footnote{\url{https://sourceforge.net/projects/germandict/}} and grammatical rules for pluralization to correctly decompose complex compounds. Since fewer resources exist for Dutch, we modified the German tool by replacing the dictionary with one from OpenOffice\footnote{\url{https://wiki.openoffice.org/wiki/Dictionaries}}, adjusting pluralization rules to fit Dutch morphology, and refining the splitting logic to avoid over-splitting, which initially led to incorrect matches.

\subsection{Lemmatization and String Matching}
To detect contentious terms in text we combine lemmatization with exact string matching. Lemmatization maps words to their canonical (dictionary) form, enabling the detection of inflected terms. 
We employed the lemmatizers from Stanford Stanza \cite{qi2020stanza}, which combine rule-based and machine-learning methods balancing precision and generalization.
Since many terms in our vocabulary are antiquated or slurs, they are underrepresented in standard training corpora.
To mitigate this, Stanza's character-level processing enables morphological analysis of words, i.e. analysing their base, inflections to express number or tense, etc., improving generalization to rare and unseen terms. 
We also leveraged Stanza’s custom rule support, implementing 162 rules to enhance accuracy for vocabulary terms that the default lemmatizer misprocessed. To enhance the tool's efficiency, all vocabulary terms were pre-processed and stored in lemmatized form.

After lemmatization, we opted for exact string matching over substring or similarity-based methods, because vocabulary terms often appear as substrings in unrelated words. Unlike domains where slurs are frequently obfuscated (e.g., numbers replacing letters), CH texts are well-formed, officially transcribed records without intentional evasion. Therefore, techniques for detecting such variations are unnecessary in our case. In the case of terms consisting of multiple tokens, the term is only matched with text spans that contain the exact same tokens (after lemmatization) in the exact same order.

\subsection{NER} 

To prevent false detections of vocabulary terms within named entities, we integrated a Named Entity Recognition (NER) module into our pipeline. One common case is terms appearing as surnames, e.g. ``Sordo'' in Italian. Using the NER module from Stanza, known for its robustness and efficiency, we ensure that terms within named entities are excluded from detection. While the LLM module could also handle this task, relying on it would impact performance due to additional processing time. The dedicated NER component allows us to achieve comparable results efficiently, maintaining both accuracy and speed.

\subsection{LLM Disambiguation}
Since certain vocabulary terms can be contentious in some contexts while neutral in others, simply flagging all detected instances would overwhelm the annotations with false positives. To address this, we implemented an LLM-based disambiguation module to assess whether a detected term is used in a contentious sense based on its surrounding context.
This task is closely related to Word Sense Disambiguation (WSD). Traditional approaches rely on BERT-like encoder models to generate contextual embeddings, which are then classified into predefined senses using either fine-tuned Transformer layers or k-Nearest Neighbors classifiers \cite{bevilacqua2021recent, loureiro2021analysis}. However, such methods require a large amount of annotated training data, which was not available for most vocabulary terms. 

A more suitable approach is the use of 
LLMs, which have demonstrated strong knowledge acquisition and contextual understanding \cite{sumanathilaka2024can}. Recent research has successfully leveraged zero-shot and few-shot prompting with LLMs for classification tasks, including WSD \cite{yae2024leveraging}, without requiring fine-tuning. This made them an attractive solution given our need for high adaptability with minimal training data. 
We tested models of varying sizes and architectures (see Sec~\ref{sec:results}), and conducted multiple rounds of prompt engineering to refine the structure and accuracy of the responses. See Appendix \ref{app:LLMs} for further information on the LLMs and the prompts. 

\section{Evaluation and Discussion} 
\label{experiments}
Our evaluation focused primarily on precision to minimize false positives (i.e., incorrectly flagged terms), which can undermine user trust \cite{Sasse2015Scaring} and acceptance, leading to rejection by users and institutions.
A major challenge in evaluating our tool was the lack of ground truth data, as CH metadata annotated for derogatory terms is scarce, if not non-existent. To overcome this, we manually created an evaluation dataset through a series of niche-sourcing campaigns, ensuring a curated and contextually rich set of records. This dataset is also made publicly available as a benchmarking resource for detecting harmful language in CH metadata. 

Using the evaluation dataset and a large set of unannotated CH metadata to assess the tool’s computational performance, we evaluated two main aspects: precision and throughput. We also examined the impact of the NER and LLM-based modules on both metrics. Additionally, we assessed precision separately for each language, testing multiple LLMs, to understand performance variations caused by differences in the vocabulary and the technical components, which sometimes do not perform consistently across all languages.

\subsection{Creation of the Evaluation Dataset}
\label{sec:eval_data}
To construct a dataset suitable for evaluating precision, we needed a representative set of records containing contentious terms. However, such terms are sparse in CH metadata, making random selection ineffective for achieving sufficient vocabulary coverage. To overcome this, we first searched for records containing vocabulary terms in a pool of 5 million Europeana records, utilizing the lemmatization and string-matching components of our pipeline. 
This allowed us to extract a subset of texts covering a wide range of vocabulary terms. For terms not found in Europeana, we manually gathered additional texts from other relevant open sources, such as Wikipedia\footnote{\url{https://www.wikipedia.org/}}, online dictionaries, and other cultural heritage digital platforms besides Europeana. In a few cases where multiple examples were unavailable for a term, we used ChatGPT\footnote{\url{https://chatgpt.com/}} to generate supplementary examples, which were subsequently reviewed by experts. Instead of replicating Europeana’s term distribution—where some terms are underrepresented—we opted for a more balanced dataset, ensuring all terms were included without over-representing the most frequent ones.
This targeted approach enabled effective precision assessment but, due to its non-random nature, did not provide meaningful insights into accuracy or recall.

To validate the detections and create a gold-standard evaluation dataset, we conducted eight niche-sourcing campaigns on CrowdHeritage~\cite{kaldeli2021crowdheritage, ralli2020crowdheritage}, a crowdsourcing platform for CH metadata enrichment. Unlike traditional crowdsourcing, niche sourcing involves domain experts, CH professionals, and affected communities as annotators. Participants could upvote or downvote detections, provide feedback, and highlight additional biased or derogatory terms that were not flagged or not included in the vocabulary.
In total, 145 participants contributed, evaluating more than 3,000 detections and casting more than 10,000 validation votes. When consensus was unclear (i.e., the absolute difference between upvotes and downvotes was less than two), campaign organizers manually reviewed these cases, considering user feedback on why certain detections were contested.

This process resulted in a curated evaluation dataset of around 3,700 texts covering all vocabulary terms. We publicly release this dataset alongside this work as a benchmarking resource for detecting contentious terms in CH metadata, complementing our vocabulary.

\subsection{Results and Discussion}
\label{sec:results}
To evaluate the tool's inference speed, we used a collection of unlabelled, de-duplicated texts from 5 million CH records (the collection referenced in~\ref{sec:eval_data} for the creation of the evaluation dataset), totaling over 690 million characters. Throughput was measured as the average characters per second processed by the tool in five runs conducted in an isolated environment. For precision, we relied on the constructed annotated evaluation dataset. All experiments were performed on a server with a single NVIDIA GeForce RTX 4090 using llama.cpp\footnote{\url{https://github.com/ggml-org/llama.cpp}}.

We first conducted an ablation study to evaluate the tool's performance with and without the NER and LLM-based modules, examining their impact on precision and throughput. For this benchmark, we used the Mixtral-8x7B model quantized to 3 bits. As shown in Table \ref{tab:throughput}, the base tool's text processing speed is very high (15,112 characters per second), with the primary bottleneck being LLM inference. This is partially alleviated by the NER module, which efficiently filters out irrelevant texts without harming precision.
Regarding precision, the NER module slightly improves accuracy when added to the base tool, though this increase depends on the presence of named entities in the evaluation dataset—a relatively rare occurrence in our case. However, in CH platforms, there are fields such as creator and contributor, which typically contain named entities. Therefore, we anticipate a more substantial impact from the NER module in real-world applications. As expected, the LLM module significantly enhances precision by filtering out many false positives caused by ambiguous terms, which would otherwise overwhelm the results with incorrect detections. 

\begin{table}[t]
    \centering
    \begin{tabular}{cc|cc}
        \toprule
        LLM & NER & Precision$\uparrow$ & Throughput$\uparrow$ \\
        \midrule
         \xmark & \xmark & 0.70 & 15112 \\
         \xmark & \ding{51} & 0.71 & 15092 \\
        \ding{51} & \xmark & 0.86 & 787 \\
        \ding{51} & \ding{51} & 0.87 & 813 \\
        \bottomrule
    \end{tabular}
    \caption{The effect of NER and LLM on performance.}
    \label{tab:throughput}
\end{table}

\begin{table*}
    \centering
    \begin{tabular}{l|ccccc|c}
        \toprule
       \multirow{2}{*}{Model} & \multicolumn{6}{c}{Precision $\uparrow$}\\
        & nl & en & fr & de & it & Aggregated \\
        \midrule
        EuroLLM-9B & 0.42 & 0.74 & \textbf{0.97} & 0.83 & 0.97 & 0.73 \\
        Llama-3.1-8B-Q8 & \textbf{0.76} & 0.83 & \textbf{0.97} & 0.88 & 0.97 & \textbf{0.88} \\
        Ministral-8B & 0.73 & \textbf{0.89} & \textbf{0.97} & 0.87 & 0.97 & \textbf{0.88}  \\
        Mistral-Small-24B-Q8 & 0.66 & \textbf{0.89} & \textbf{0.97} & 0.87 & \textbf{0.98} & 0.86 \\
        Mixtral-8x7B-Q3 & 0.72 & 0.83 & \textbf{0.97} & \textbf{0.89} & \textbf{0.98} & 0.87 \\
        StableLM-2-1.6B & 0.60 & 0.72 & 0.96 & 0.82 & 0.95 & 0.80 \\
        StableLM-2-12B & 0.72 & 0.82 & 0.96 & 0.87 & 0.97 & 0.86  \\
        \bottomrule
    \end{tabular}
    \caption{Precision of the tool, utilizing different LLMs. Bold indicates the highest precision per language, and aggregated.}
    \label{tab:precision}
\end{table*}

We also evaluated a variety of LLMs across all languages, with the results presented in Table \ref{tab:precision}. Due to memory constraints, some LLMs were quantized, indicated in the table by the letter `Q' followed by the number of bits. Precision remained consistently high for Italian and French, which we attribute to the low percentage of ambiguous terms for these languages in the vocabulary. In contrast, precision was lower for Dutch, primarily due to the high number of ambiguous Dutch terms combined with the lower performance of LLMs on Dutch—likely a result of limited Dutch data used in LLM pretraining. Overall, Llama-3.1-8B and Ministral-8B achieved the best precision, with Mixtral-8x7B and StableLM-2-12B performing comparably well. In terms of throughput, StableLM-2-1.6B was the fastest model, with Llama-3.1-8B coming second, providing a good trade-off between speed and accuracy. 

\section{Application \& Impact}
To maximize the usability and impact of our tool, we developed a public API endpoint that allows integration into other platforms and services. This API is the basis for a stand-alone web application and it has been deployed in two large-scale CH platforms, Europeana and MINT.

\subsection{Standalone Web Application}
To ensure accessibility for users with little to no technical expertise—an important request from many practitioners in the CH domain—we developed a standalone web application\footnote{\url{https://debias-tool.ails.ece.ntua.gr/}} with an intuitive and interactive interface. The web app provides two main functionalities: (1) users can input custom texts and directly view the detected terms alongside contextual explanations from the vocabulary, detailing why each term is considered derogatory; (2) users can analyze larger datasets by uploading a ZIP file containing multiple text files, with the option to receive via email a JSON file with machine-readable annotations and a statistical report of the results.
Users can customize the service by (de-)activating the components for NER and LLM-based disambiguation. This enables them to experiment and optimize performance based on their specific needs--prioritizing processing speed or precision.
Independently from such practical usage, the application also serves as a key resource for showcasing the tool and engaging with a broader audience about the issues of bias in CH metadata. In earlier workshops, this enabled us to gather feedback that was useful to refine earlier versions of the tool and the vocabulary based on real-world cases.

\subsection{Integration with External Platforms}
\textbf{Europeana}, a platform providing access to digital heritage across Europe with contributions from over 3,200 CHIs and more than 10,000 daily visits on europeana.eu, has integrated our tool into its metadata ingestion pipeline. This integration allows data providers to analyze their records for contentious terms before publication. Additionally, Europeana has developed a UI component that highlights these terms on item pages, offering contextual explanations sourced from our vocabulary and linking to the terms published in the EU Vocabularies platform for further reference.
This functionality can display detections for newly ingested metadata while also surfacing detections from our tool’s application on over 7.9 million legacy metadata records, resulting in more than 77,000 annotations. This large-scale deployment enhances transparency by identifying and contextualizing contentious terms across the platform.

\textbf{MINT}, the Metadata Interoperability Tool, is a service that enables organizations with diverse data formats to transform and map their metadata into a standardized schema, which can then be exported to an online repository for harvesting by consuming platforms. Currently, MINT facilitates the publication of over 5.2 million CH records from approximately 200 organizations to multiple platforms, including Europeana.
Our tool has been integrated into MINT, allowing users to detect contentious terms in their datasets and annotate them. Organizations can analyze their data in an easy and intuitive way via the MINT UI, without requiring technical expertise to interact with an API. Importantly, this integration gives CHIs direct access to the tool's functionalities within MINT, while also retaining full control over how to handle the results, ensuring they can assess and integrate the tool’s insights according to their specific needs and policies.

\section{Conclusions}
In this work, we presented a comprehensive approach to detecting and contextualizing harmful language in CH metadata. We developed a rich multilingual vocabulary of contentious terms, integrating it with an AI-based tool designed to identify potentially harmful language while providing contextual explanations. Additionally, we curated a niche-sourced evaluation dataset, enabling a grounded assessment of our tool's precision across multiple languages. This dataset is publicly available, supporting further research on harmful language detection in CH contexts. Our tool was deployed at scale via a standalone web app, and integrated with major CH platforms. 

Future work will focus on expanding the vocabulary to include more terms and extending coverage to additional languages, enhancing its applicability across diverse cultural contexts. We also plan to investigate advanced technical solutions to improve the precision of detection, including the integration of more sophisticated NLP techniques and adaptive models that better capture context and nuanced meanings. These advancements will also aim to increase accuracy while maintaining transparency, ultimately contributing to more inclusive and representative digital heritage platforms.

\section*{Limitations}
One limitation of our approach emerged during our community discussions. Participants pointed out a persistent bias in the form of information omissions, leading to incomplete representations of their culture and history. This issue was particularly evident in collections related to gender and sexual identity (e.g., excluding information about LGBTQIA+ relationships) and migration and colonial histories (e.g., using inaccurate or outdated vocabulary to describe cultural activities or circumstances). Often, these CH objects were described from an uninformed and biased perspective, outside the community. Similarly, a general lack of adequate and representative language in the past makes these collections less accessible. This specific form of contentiousness cannot be fully addressed by a tool built for detecting terms in the metadata. 

Another limitation is that, although we adopted an inclusive approach to co-create the vocabulary with marginalized communities, we could not apply this process to the entire vocabulary. Our decision to focus the community work on specific themes and languages meant that we were unable to validate and discuss the vocabulary terms for all themes across every language. However, this was the only feasible option due to time and budget constraints. A full validation across all languages and by all affected communities would be a lengthy process, requiring substantial resources.

Finally, a third limitation pertains to the technical approach. The tool we provide as open source has high computational requirements, necessitating at least a mid-range GPU, which may limit accessibility for some who lack access to such resources. To address this, we make the API endpoint and the web application publicly available for anyone to use, with the goal of increasing accessibility. On the other hand, computational constraints prevented us from experimenting with larger models that could have potentially enhanced the tool's performance. However, we do not view this as a significant limitation, as our model has demonstrated strong performance even with limited resources. By making the project open source, we enable teams with the necessary resources to build upon it and take it to the next level.

\section*{Ethical Considerations}
\label{sec:ethical}
This work was conducted in the context of an EU-funded initiative and adhered to the ethical standards set by the European Commission. As part of the funding requirements, the project underwent an independent ethics review prior to the start of the study, which concluded that the research was in line with ethics and security requirements. An external advisory board of independent experts was appointed to provide ongoing oversight and guidance, particularly in relation to the use of potentially sensitive or harmful language. Their recommendations were regularly sought and integrated into the research process. All activities were also carried out in accordance with the institutional ethical guidelines of the partner organizations involved.

To support contributors involved in the vocabulary curation,  community engagement, and detection evaluation activities, all participants were informed in advance about the sensitive nature of the content and were provided with an introductory session, inclusive guidelines, and content warnings. They were free to withdraw from any activity at any time without consequence. Remuneration and support for the community engagement were discussed and agreed upon with each group, reflecting local norms and the preferences of the participating communities. In some cases, such as in Belgium and the Netherlands, individual participants were directly compensated for their time and input. In others, including the UK and Germany, the available budget was directed toward the community more broadly—either by employing a community member over a longer period or compensating a workshop organizer from within the community. Indicative numbers include €200-€300 for individual participants and €750-€800 for workshop organizers, or €4,320 for a six-month community liaison role. These figures reflect what was feasible within the constraints of our budget and were always stipulated in mutual agreement with contributors. We believe that fair compensation is an important principle in collaborative research, especially when working with communities on sensitive topics. At the same time, we recognise that not all works will have the same resources. Transparency with contributors about available financial support is essential, and we provide these amounts for reference rather than as prescriptive standards for future work.

There are also certain ethical considerations regarding the technical developments and the use of resources created in this work. In our vocabulary, we provide suggested alternative terms for some vocabulary entries. When discussing our work with the CH community, a key concern was avoiding the simple replacement of contentious terms or ``sanitizing'' the metadata. While this was not our approach, we discussed the benefits of augmenting or enriching descriptions with those alternatives as this could offer the possibility to find CH objects when searching for these more appropriate alternative terms, improving accessibility. However, this approach was not pursued by us. It would require additional research since the applicability of the suggested terms depends on context, much like the contentiousness of many terms.

Another important ethical consideration is the potential harm of deploying an underprepared or low-performing harmful language detection tool. Flooding a platform with false positives could significantly erode user trust in the system, as users may begin to question its reliability and accuracy. Moreover, false flagging may unintentionally diminish the significance of the flagged issues in users' minds. When contentious terms are flagged without proper context or accuracy, it risks trivializing important conversations and reducing awareness of these issues. Therefore, ensuring the tool is well-calibrated and accurate is crucial to maintaining both its credibility and its effectiveness in addressing harmful language.

\section*{Acknowledgements}
This work has been partially supported by the project Detecting and Cur(at)ing Harmful Language in Cultural Heritage Collections (DE-BIAS), Grant No. 101100744, under the EU programme DIGITAL-2022-CULTURAL-02. It would not have been possible without the collaborative effort and dedication of all project partners. We are also grateful to the individuals and community members who participated in workshops, curated vocabulary terms, organized events, provided invaluable feedback, tested the tool, and supported this work in numerous ways. Their time, insights, trust and commitment to fostering more inclusive and reflective cultural heritage collections have been instrumental. 
The authors also gratefully acknowledge financial support from the Research Committee of the National Technical University of Athens (NTUA) through a scholarship.

\bibliography{custom}
\bibliographystyle{acl_natbib}

\appendix

\section{Vocabulary Examples}
\label{app:voc_example}
In Table~\ref{tab:voc_example} we present an example of a vocabulary entry. 
\begin{table*}

    \centering
    \begin{tabular}{p{2cm}p{6cm}p{3cm}p{1.5cm}p{1.5cm}}
        \toprule
        Contentious Term & Contentious Issue & Suggestion Note & Suggested Term & Category\\
        \midrule
        Caucasian & The term ``Caucasian'' was introduced in the late 18th century by German anthropologist Johann Friedrich Blumenbach, who idealized the Caucasus region as the origin of the “white race” and an aesthetic ideal. In the 19th century, it became a racial category encompassing Europeans, some North Africans, and parts of Western Asia, as part of pseudoscientific theories that hierarchically classified humans based on physical traits. Though later incorporated into Nazi racial ideology, the term was less central than narrower categories like ``Aryan''. In the U.S., ``Caucasian'' became a common term for describing white populations in legal, demographic, and social contexts, often as a synonym for ``white''. but it is increasingly criticized for its imprecision and roots in outdated racial theories & Use with caution in the context of people. The term is unproblematic when it refers specifically to people from the Caucasus region. & White & Ethnicity, Race\\
        \bottomrule
    \end{tabular}
    \caption{A sample entry from the vocabulary.}
    \label{tab:voc_example}
\end{table*}


\section{LLM sources and prompts}
\label{app:LLMs}
All LLMs used were found on Huggingface\footnote{\url{https://huggingface.co/}}. We used LLMs fine-tuned for instruction following, with the exception of the StableLM models which were only available as fine-tuned for chatting. We used model weights published in the GGUF format\footnote{\url{https://huggingface.co/docs/hub/en/gguf}} since that is compatible with llama.cpp. We provide URLs for the LLMs we used as footnotes: EuroLLM-9B\footnote{\url{https://huggingface.co/bartowski/EuroLLM-9B-Instruct-GGUF}} \cite{martins2024eurollm}, Llama-3.1-8B\footnote{\url{https://huggingface.co/bartowski/Meta-Llama-3.1-8B-Instruct-GGUF}} \cite{dubey2024llama}, Ministral-8B\footnote{\url{https://huggingface.co/bartowski/Ministral-8B-Instruct-2410-GGUF}}, Mistral-Small-24B\footnote{\url{https://huggingface.co/mistralai/Mistral-Small-24B-Instruct-2501}}, Mixtral-8x7B\footnote{\url{https://huggingface.co/TheBloke/Mixtral-8x7B-Instruct-v0.1-GGUF}} \cite{jiang2024mixtral}, StableLM-2-1.6B\footnote{\url{https://huggingface.co/mradermacher/stablelm-2-1_6b-chat-i1-GGUF}} \cite{bellagente2024stable}, StableLM-2-12B\footnote{\url{https://huggingface.co/stabilityai/stablelm-2-12b-chat-GGUF}}.

We produced a different prompt format for each language, starting with a base one in English and translating it to the other languages. In the preliminary stages, we tested using the same English prompt for all languages and simply denoting the language of the text, but this produced sub-par results. We provide as examples the English and Italian prompts we used for Llama-3.1-8B, as well as the English prompt for Mixtral8x7B in Table \ref{tab:prompts}. The prompts used for different LLMs mostly differ on the special tokens that each LLM uses to separate system and user prompts. 

One important aspect of the prompts is steering the output of the model towards an easily extractable answer. This was a problem we faced early on during our prompt engineering process, which we mitigated by a very structured format and by explicitly specifying the desired outputs. This very simplistic output format does not provide the model with enough tokens to ``think'' through the correct answer. We experimented with longer formats, including Chain-of-Thought, but they severely reduced the throughput at a level that was not acceptable for a tool meant to process millions of records. Nevertheless, in future versions of the tool we could provide longer prompt formats to users who do not prioritize throughput.

\begin{table*}[]
    \centering
    \begin{tabular}{p{3.5cm}p{10.5cm}}
        \toprule
        Model/Language & Prompt \\
        \midrule
        Llama-3.1/English & \texttt{<|begin\_of\_text|><|start\_header\_id|>system<|end\_header \_id|> The term "\{term\}" can be contentious when used in some contexts. Here is an explanation of why "\{term\}" can be considered contentious: \{vocabulary \_context\}<|eot\_id|><|start\_header\_id|>user<|end \_header \_id|>\textbackslash nIs "\{term\}" used in a contentious manner in the following text? Answer with a simple yes or no.\textbackslash n\textbackslash nText:\textbackslash n\{text\}<|eot\_id|><|start\_header\_id|> assistant <|end \_header \_id|>} \\
        Llama-3.1/Italian & \texttt{<|begin\_of\_text|><|start\_header\_id|>system<|end\_header \_id|> Il termine "\{term\}" può essere controverso quando utilizzato in alcuni contesti. Ecco una spiegazione del perché "\{term\}" può essere considerato controverso: \{vocabulary\_context\} <|eot\_id|><|start \_header\_id|>user<|end\_header\_id|>\textbackslash nIl termine  "\{term\}" è utilizzato in modo controverso nel seguente testo? Rispondi con un semplice sì o no.\textbackslash n\textbackslash nTesto:\textbackslash n\{text\} <|eot\_id|><|start\_header\_id|>assistant<|end\_header\_id|>} \\
        Mixtral8x7B/English & \texttt{<s> [INST] The term "\{term\}" can be contentious when used in some contexts. Here is an explanation of why "\{term\}" can be considered contentious: \{vocabulary\_context\} \textbackslash n\textbackslash nQuestion:\textbackslash nIs "\{term\}" used in a contentious manner in the following text? Answer with a simple yes or no.\textbackslash n\textbackslash nText:\textbackslash n\{text\}\textbackslash n\textbackslash nAnswer:\textbackslash n[/INST]\textbackslash n} \\
        \bottomrule

    \end{tabular}
    \caption{Some of the prompts used for different LLMs and languages.}
    \label{tab:prompts}
\end{table*}

\end{document}